\pdfoutput=1

\documentclass[11pt]{article}

\usepackage[final]{acl}

\usepackage{times}
\usepackage{latexsym}
\usepackage{amsmath}
\usepackage{amssymb}
\usepackage{booktabs}
\usepackage{multirow} 
\usepackage{bm}
\usepackage{enumitem}

\usepackage[T1]{fontenc}

\usepackage[utf8]{inputenc}

\usepackage{microtype}

\usepackage{inconsolata}

\usepackage{graphicx}

\title{Entropy-based Coarse and Compressed Semantic \\ Speech Representation Learning}

\author{
Jialong Zuo$^{1}$, Guangyan Zhang$^{2}$, Minghui Fang$^{1}$, Shengpeng Ji$^{1}$, Xiaoqi Jiao$^{2}$ \\ \textbf{Jingyu Li$^{2}$, Yiwen Guo$^{3}$, Zhou Zhao$^{1}$\thanks{Corresponding author.}} \\
$^{1}$Zhejiang University\hspace{1em}  $^{2}$LIGHTSPEED\hspace{1em} 
$^{3}$Independent Researcher\\
\href{mailto:jialongzuo@zju.edu.cn}{\texttt{jialongzuo@zju.edu.cn}} \hspace{1em} 
\href{mailto:zhaozhou@zju.edu.cn}{\texttt{zhaozhou@zju.edu.cn}}
}

\begin{document}
\maketitle
\begin{abstract}
Discrete speech representation learning has recently attracted increasing interest in both acoustic and semantic modeling. Existing approaches typically encode 16 kHz waveforms into discrete tokens at a rate of 25–50 tokens per second. However, given that speech generally conveys only 2–5 words per second, such fine-grained tokenization introduces redundancy and hinders efficiency in downstream training and inference. Moreover, semantic speech representations at this frequency primarily capture phonetic-level information, while semantic understanding may not require such detailed token-level resolution. To address these limitations, we propose an entropy-based dynamic aggregation framework for learning compressed semantic speech representations. A speech language model is first pre-trained via next-token prediction on large-scale unlabeled data to capture frequent token patterns. Predictive entropy is then used to adaptively determine aggregation boundaries, followed by a cross-attention module that fuses information within each segment. By adjusting the entropy threshold, the granularity and compression ratio of the representations can be flexibly controlled. Experiments on ASR, speech-to-text translation, and voice conversion tasks demonstrate that the compressed representations perform on par with or better than dense token sequences, demonstrating the effectiveness of the proposed approach.
\end{abstract}

\section{Introduction}
In recent years, discrete representation learning of speech \cite{encodec,dac,zhang2023speechtokenizer,ji2024wavtokenizer} has achieved remarkable success and has gradually emerged as a crucial bridge between the speech modality and large language models \cite{xie2024mini1,xie2024mini2,xu2025qwen2}. A notable direction within this paradigm is semantic-level discretization, aimed at retaining only the high-level semantic content of audio signals, enabling much lower bitrates. For tasks that primarily rely on speech content such as automatic speech recognition (ASR) or speech-to-text translation, semantic tokens provide a more efficient and task-relevant representation.

Self-supervised learning (SSL) of speech \cite{schneider2019wav2vec,baevski2020wav2vec,hubert,chung2021w2v,chen2022wavlm} has demonstrated remarkable success in capturing rich and transferable speech representations across a wide range of downstream tasks. Building on this foundation, a growing number of studies have sought to extract semantic units from SSL-derived speech features. HuBERT \cite{hubert} represents a pioneering effort in this direction, employing k-means clustering to tokenize speech, where the resulting discrete units are used as training targets for the speech encoder and are found to exhibit strong phoneme-level correlations \cite{choi2024self}. Subsequent works have adopted similar approaches to derive semantic representations: Spirit-LM \cite{nguyen2025spirit} extracts HuBERT features and applies k-means clustering with 500 centroids to define its basic semantic units, while Vevo \cite{zhang2025vevo} follows a comparable methodology. SpeechTokenizer \cite{zhang2023speechtokenizer} leverages semantic representations extracted from a HuBERT-based teacher model to guide the distillation of the first layer of residual vector quantization (RVQ), thereby generating discrete units intended to capture semantic content. Similarly, AudioLM \cite{borsos2023audiolm} utilizes semantic tokens derived from w2v-BERT \cite{chung2021w2v} to encode high-level semantic information from audio inputs.

In addition to extracting semantic units from self-supervised speech representations, several methods employ supervised approaches such as ASR or phoneme recognition to obtain semantic tokens. For instance, CosyVoice \cite{du2024cosyvoice} proposes to use a supervised ASR module to construct a supervised semantic speech (S3) tokenizer. FACodec \cite{ju2024naturalspeech} directly leverages phoneme labels (specifically, frame-level phoneme annotations obtained from internal alignment tools) to train discrete content representations in a supervised manner. While the effectiveness of these semantic tokenizers in capturing high-level semantic units has been validated across various downstream tasks such as ASR \cite{yang2023universalspeechdiscretetokens}, text-to-speech (TTS) \cite{du2024cosyvoice,wang2024maskgct}, and voice conversion (VC) \cite{kim2023unitspeech,liu2024zero,zuo2025enhancing}, several critical limitations remain to be addressed. First, existing semantic discretization approaches typically compress speech into 25 or more tokens per second, meaning that long speech utterances often result in an excessive number of discrete units. This introduces a substantial computational burden for training and inference in downstream models. Second, the semantic tokens often encode excessive phonetic detail \cite{sicherman2023analysing,cho2023evidence}, which may be unnecessarily fine-grained for tasks focused on semantic understanding. Therefore, both further compression and coarser-grained semantic representations are necessary and desirable.

In this study, we propose an entropy-based semantic token compression strategy for dynamically extracting semantic speech representations at varying levels of granularity. Specifically, building upon HuBERT units (discretized via a K-Means clustering method), we achieve further sequence compression by training a lightweight auto-regressive language model (LM) on large-scale pretrained speech data and computing the entropy of the next-token distribution. By leveraging the LM’s ability to model frequently occurring patterns (i.e., sequences of adjacent tokens), we define an entropy threshold to segment the token stream and merge adjacent tokens accordingly. This approach effectively reduces the sequence length. Furthermore, we demonstrate that modifying the entropy threshold and segmentation criteria enables flexible control over compression granularity, which subsequently influence the performance of downstream semantic understanding and generation tasks. The contributions of this work are summarized as follows:
\begin{itemize}[itemsep=1pt]
    \item We propose an entropy-based token aggregation framework that adaptively compresses semantic speech representations by merging adjacent tokens based on predictive uncertainty derived from a lightweight autoregressive language model.
    \item Our method enables controllable granularity in semantic token sequences, significantly reducing sequence length while retaining essential semantic content, thus enhancing efficiency for downstream speech understanding and generation tasks.
    \item Empirical results across multiple benchmarks demonstrate that our approach achieves a favorable balance between compression rate and task performance, validating its effectiveness and scalability.
\end{itemize}

\section{Related Work}
\subsection{Self-supervised Speech Representation Learning}
Self-supervised learning (SSL) has emerged as a powerful paradigm for extracting rich and generalizable representations from large-scale unlabeled speech corpora. Recent advances such as wav2vec \cite{baevski2020wav2vec}, HuBERT \cite{hubert}, and WavLM \cite{chen2022wavlm} have demonstrated the effectiveness of SSL in capturing both phonetic and semantic information from raw audio signals. These models typically rely on pretext tasks such as masked prediction or contrastive learning to model the temporal structure of speech without requiring manual annotations. For example, wav2vec employs a contrastive loss to distinguish true audio segments from distractors, thereby learning high-level representations. HuBERT adopts a masked prediction objective, where targets are derived from offline k-means clustering, encouraging the model to learn context-aware representations. W2v-BERT \cite{chung2021w2v} further combines contrastive and masked learning in an end-to-end framework to leverage the benefits of both objectives. Meanwhile, WavLM further introduces a denoising objective to enhance robustness against noise and speaker variation.

\begin{figure*}[ht]
\centering
\includegraphics[height=6.5cm, width=16cm]{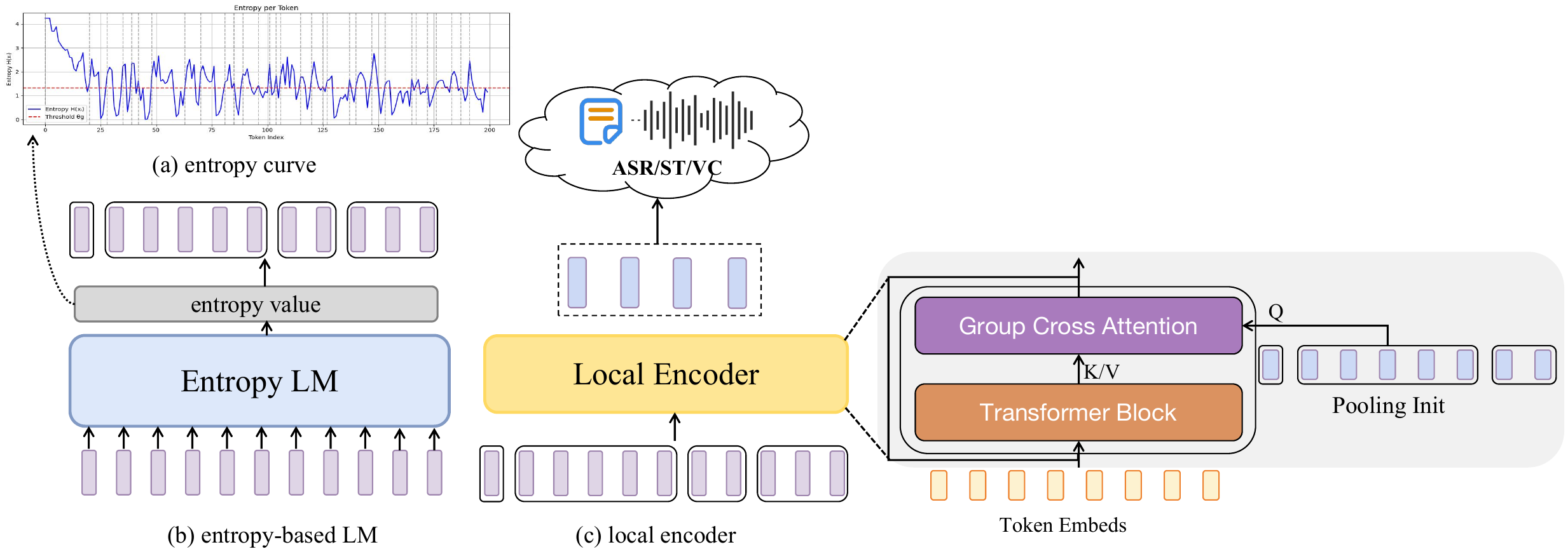}

\caption{Illustration of the proposed entropy-based semantic token compression framework. Figure (a) shows the next-token entropy curve predicted by an autoregressive language model trained on HuBERT units, where a designed entropy threshold is used to segment token groups. Figure (b) illustrates the entropy language model, which is trained to estimate the conditional entropy of the next token given preceding context, and whose predictions are used to determine grouping boundaries. Figure (c) presents the local encoder, which incorporates the compressed token groups via a cross-attention mechanism for downstream understanding and generation tasks.}
\label{entropy_overall}
\end{figure*}

These models typically operate at a fixed frame rate (e.g., 50 Hz), producing fine-grained representations that align closely with sub-phonemic units \cite{choi2024self}. While effective for phonetic modeling, such granularity can be redundant for tasks requiring only high-level semantic understanding, motivating the need for more compact and task-relevant representations.
\subsection{Extracting Semantic Units from Speech} Semantic units, or semantic tokens, are discrete representations derived from speech that encapsulate higher-level phonetic or linguistic information. These units are typically extracted either from self-supervised learning (SSL) models or through supervised training paradigms.

In SSL-based approaches, semantic tokens are often obtained by quantizing hidden representations using methods such as K-means or VQ-VAE \cite{vqvae}. This process, applied to pretrained models like HuBERT or Wav2Vec 2.0, yields discrete indices that correlate with phonetic or semantic properties. Some SSL models incorporate internal quantizers, trained jointly with the backbone network, producing token-like outputs that serve as training targets. While conventionally termed "semantic", recent findings suggest that these tokens may primarily encode phonetic structure rather than higher-level semantics \cite{cho2024self,cho2023evidence,choi2024self}. Nevertheless, they have demonstrated strong performance across a range of downstream tasks.

Supervised methods, by contrast, offer a more explicit mechanism for semantic token extraction. Models such as the S³ Tokenizer \cite{du2024cosyvoice} integrate vector quantization within a Transformer-based architecture and are trained with ASR-like objectives on large-scale paired speech-text corpora. This allows for more preservation of paralinguistic information
than directly transcribing speech into text. These supervised tokenizers are trained on massive paired speech-text data, and have demonstrated rich speech content understanding capabilities.

\section{Method}
\subsection{Overall Architecture}
The overall architecture of the proposed entropy-based semantic compression framework is illustrated in Figure \ref{entropy_overall}. It consists of three main components: (1) Discrete Semantic Token Extraction: input speech is first processed by a pretrained HuBERT model, and the resulting continuous representations are discretized using k-means clustering to obtain frame-level semantic tokens. (2) Entropy-Based Grouping Module: a lightweight autoregressive language model is trained on these tokens to model their distribution and compute token-wise conditional entropy. Based on a designed entropy threshold, adjacent tokens with low predictive uncertainty are merged into variable-length groups, enabling sequence compression. (3) Cross-Attentive Local Encoder: the compressed token groups are fed into a local encoder equipped with a cross-attention mechanism, enabling efficient incorporation of compressed representations into downstream understanding and generation tasks. The framework allows flexible control over the compression ratio by adjusting the entropy threshold, supporting different granularity levels for diverse applications.

\subsection{Discrete Semantic Token Extraction}
An input waveform is first encoded by a pretrained HuBERT model and discretized via k-means clustering into frame-level semantic tokens with a codebook size of $K$:
\[
\mathbf{u} = \{u_1, u_2, \dots, u_N\}, \quad u_i \in \{1,\dots,K\}.
\]
These tokens serve as the discrete input units for subsequent entropy-based compression model.
\subsection{Entropy-Based Grouping Module}

The primary objective of the Entropy-Based Grouping Module is to adaptively compress the token sequence $\mathbf{u}$ by leveraging entropy as a measure of uncertainty in the token predictions. To achieve this, we train a lightweight autoregressive language model (LM) on the discrete token sequence. The LM predicts the probability distribution of each token $u_i$ conditioned on its preceding tokens, thereby capturing the sequential dependencies and uncertainties in the token sequence. Formally, for each position $i$, the LM estimates the conditional distribution:

\[
 p(u_i \mid u_{<i}) \quad \text{for } u_i \in \{1, \dots, K\},
\]

where $u_i$ represents the current token and $u_{<i}$ denotes the sequence of preceding tokens. This distribution is essential for quantifying the uncertainty associated with each token prediction.

\noindent\textbf{Token-wise Entropy.}
To measure the uncertainty of each token prediction, we compute the token-wise entropy $H(u_i)$, which is based on the predictive distribution $p(u_i \mid u_{<i})$. This entropy quantifies the unpredictability or uncertainty in the token's prediction:

\begin{equation}
H(u_i) = -\sum_{v=1}^{K} p(u_i = v \mid u_{<i})\,\log p(u_i = v \mid u_{<i}),
\label{eq:entropy}
\end{equation}

where $H(u_i)$ is the entropy of the predicted distribution over all possible tokens $v \in \{1, \dots, K\}$. A higher entropy value indicates greater uncertainty in the prediction, while a lower entropy suggests a more confident prediction. This entropy-based measure enables an adaptive segmentation strategy, where boundaries are determined by detecting sharp increases in conditional entropy, allowing segments of low uncertainty to be merged while capturing transitions in the underlying linguistic organization.


\noindent\textbf{Boundary Identification.}
We introduce two entropy-based criteria to identify the boundaries that separate different groups of tokens. These criteria rely on both the token's individual entropy and the relative change in entropy between consecutive tokens. The first criterion uses a global entropy threshold $\theta_g$ to identify regions with high uncertainty:

\[
H(u_i) > \theta_g, \label{eq:global_threshold}
\]

This global threshold $\theta_g$ is empirically determined and ensures that tokens with sufficiently high entropy are considered as potential group boundaries. The second criterion captures abrupt changes in uncertainty between adjacent tokens by examining the relative difference in entropy between $u_i$ and $u_{i-1}$:

\[
H(u_i) - H(u_{i-1}) > \theta_r, \label{eq:relative_threshold}
\]

where $\theta_r$ is a relative threshold that emphasizes significant increases in uncertainty. By setting this criterion, we can identify boundaries that reflect changes in the predictability of the token sequence, which likely correspond to boundaries at the phonetic or sub-phonemic level, consistent with the characteristics of HuBERT tokens.

\noindent\textbf{Group Formation.}
Once the boundary conditions are applied, we identify the positions $b_0, b_1, \dots, b_M$ where the boundaries occur, with $b_0 = 0$ and $b_M = N$, where $N$ is the total number of tokens in the sequence. These indices define the boundaries of $M$ contiguous token groups:

\[
 g_j = \{u_{b_{j-1}+1}, \dots, u_{b_j}\}, \quad j = 1, \dots, M.
\]

Each group $g_j$ consists of tokens between two consecutive boundary indices. This adaptive grouping process dynamically adjusts the length of each group based on local token predictability. In particular, the groups formed in regions with high uncertainty will likely be shorter, while groups in more predictable regions will be longer. The final result is a compressed sequence $\{g_j\}_{j=1}^M$, where each group represents a semantically coherent subset of the original token sequence. This compression helps reduce the overall sequence length while maintaining key semantic information.

By tuning the thresholds $\theta_g$ and $\theta_r$, the Entropy-Based Grouping Module offers fine-grained control over the trade-off between compression ratio and the preservation of semantic structure. This adaptive grouping process plays a critical role in the overall compression framework, ensuring that high-entropy regions are appropriately compressed while minimizing the loss of important semantic content.

\subsection{Cross-Attentive Local Encoder}

The cross-attentive local encoder aims to derive compact group-level representations $\{\mathbf{p}_j\}$ from token sequences grouped into segments $\{g_j\}$. Unlike standard pooling approaches, our design iteratively refines group embeddings through stacked cross-attention layers, allowing information aggregation from the constituent tokens in each group.

\noindent\textbf{Token Embedding.}
Each discrete token $u_i$ is embedded into a continuous vector $\mathbf{e}_i \in \mathbb{R}^d$ via a learnable embedding matrix $\mathbf{W}_e \in \mathbb{R}^{K \times d}$, where $K$ is the vocabulary size and $d$ is the embedding dimension. The input token sequence is thus transformed into $\{\mathbf{e}_i\}_{i=1}^{N}$.

\noindent\textbf{Initialization of Group Queries.}
For each group $g_j = \{u_{b_{j-1}+1}, \dots, u_{b_j}\}$, we initialize its representation $\mathbf{p}_j^{(0)} \in \mathbb{R}^{d}$ by applying max pooling over the token embeddings:
\[
\mathbf{p}_j^{(0)} = \mathrm{MaxPool}\left( \left\{ \mathbf{e}_i \mid i \in g_j \right\} \right).
\]

\noindent\textbf{Cross-Attention Update.}  
We employ $L$ cross-attention layers to iteratively refine the group representations. At the $\ell$-th layer ($\ell = 1,\dots,L$), the group representation $\mathbf{p}_j^{(\ell)}$ is updated via attention over the token representations $\{\mathbf{h}_i^{(\ell-1)}\}$, where $\mathbf{h}_i^{(0)} = \mathbf{e}_i$. The update is defined as:
\begin{align*}
\mathbf{q}_j^{(\ell)} &= \mathrm{LayerNorm} \left( \mathbf{W}_Q \mathbf{p}_j^{(\ell-1)} \right), \\
\mathbf{k}_i^{(\ell)} &= \mathrm{LayerNorm} \left( \mathbf{W}_K \mathbf{h}_i^{(\ell-1)} \right), \\
\mathbf{v}_i^{(\ell)} &= \mathrm{LayerNorm} \left( \mathbf{W}_V \mathbf{h}_i^{(\ell-1)} \right), \\
\alpha_{j,i}^{(\ell)} &= \frac{\exp\left( \mathbf{q}_j^{(\ell)\top} \mathbf{k}_i^{(\ell)} / \sqrt{d} \right)}{\sum\limits_{k = b_{j-1}+1}^{b_j} \exp\left( \mathbf{q}_j^{(\ell)\top} \mathbf{k}_k^{(\ell)} / \sqrt{d} \right)}, \\
\mathbf{z}_j^{(\ell)} &= \sum_{i = b_{j-1}+1}^{b_j} \alpha_{j,i}^{(\ell)} \mathbf{v}_i^{(\ell)}, \\
\mathbf{p}_j^{(\ell)} &= \mathbf{p}_j^{(\ell-1)} + \mathbf{W}_O \mathbf{z}_j^{(\ell)}.
\end{align*}

After $L$ cross-attention layers, the final group embedding is defined as $\mathbf{p}_j = \mathbf{p}_j^{(L)}$. Attention is inherently restricted to tokens within each group by their index range. At each layer, the query $\mathbf{q}_j^{(\ell)}$ is recomputed from the previous group embedding, facilitating hierarchical and localized aggregation of token information. Residual connections and pre-layer normalization are adopted for stable optimization. We omit positional encodings, since intra-group order is implicitly captured through attention weights.

\noindent\textbf{Group-Level Output Representation.}
The resulting sequence of group embeddings $\{\mathbf{p}_j\}_{j=1}^M$ is subsequently passed to downstream modules for further semantic understanding or generation tasks. By adjusting the parameters $\theta_g$ and $\theta_r$, the framework provides fine-grained control over the trade-off between the compression ratio and the preservation of semantic accuracy.

\section{Experiments}
\subsection{Experiment Setup}
\textbf{Datasets} We utilize the English portion of the Multilingual LibriSpeech (MLS) dataset \cite{pratap2020mls}, comprising approximately 20,000 hours of speech, to pretrain the Entropy LLM and to provide training data for both the ASR and voice conversion tasks. For ASR and zero-shot voice conversion evaluation, we use the standard test-clean subset from LibriSpeech \cite{panayotov2015librispeech}, covering a diverse set of speakers. To assess speech-to-text translation performance, we adopt the benchmark CVSS-C dataset \cite{jia2022cvss}, which is derived from the CoVoST 2 speech-to-text translation corpus \cite{wang2021covost}. We conduct experiments on three language pairs: English-Chinese (EN-CN), English-Spanish (EN-ES), and English-French (EN-FR) covering a range of widely used languages with varying linguistic characteristics.
\subsection{Training and Inference Setup}

\textbf{Entropy LLM Pretraining} The Entropy LLM follows a lightweight LLaMA-style \cite{llama} architecture with approximately 0.1 billion parameters. It consists of 8 Transformer layers with a model dimension of 1024, 16 attention heads, and a maximum sequence length of 2048. Rotary positional embeddings (RoPE) \cite{su2023roformerenhancedtransformerrotary} are used, and all hidden dimensions are aligned to multiples of 256. The model is trained for 750k steps on the English subset of MLS (20k hours) using the AdamW optimizer with a learning rate of $3 \times 10^{-4}$, weight decay of 0.01, $\beta_1 = 0.9$, $\beta_2 = 0.999$, and $\epsilon = 10^{-5}$. A cosine learning rate schedule with 2k warmup steps and a final learning rate ratio of 0.1 is applied. We utilize the open-source pretrained HuBERT model\footnote{\url{https://github.com/facebookresearch/fairseq/blob/main/examples/hubert}} and a $k$-means clustering model for semantic unit extraction, with the number of clusters set to $k = 500$. For the high compression setting at 7 Hz token rate, the global entropy threshold is $\theta_g = 9.7$.

\textbf{ASR and Speech-to-Text Translation} For ASR and speech translation tasks, we adopt an encoder-decoder architecture similar to Whisper \cite{whisper}. The encoder is configured either as a standard Conformer encoder \cite{conformer} for raw semantic units, or as a lightweight local encoder tailored for compressed coarse semantic speech representations. The decoder generates output text autoregressively. Speech translation is evaluated on the CVSS-C benchmark, covering English–Chinese, English–Spanish, and English–French pairs.

\textbf{Voice Conversion} The voice conversion model is based on a Diffusion Transformer similar to~\cite{liu2024zero}. It takes as input either original semantic tokens or compressed semantic representations, along with a reference mel-spectrogram. The output mel-spectrogram is synthesized using a pretrained HiFi-GAN vocoder \cite{hifigan}.

\begin{table*}[h]
\small
\centering
\caption{Performance comparison under different entropy thresholds for ASR, ST, and VC tasks. The entropy thresholds correspond to token rates of 24 Hz (Fine-grained), 15 Hz (Moderately compressed), and 7 Hz (Coarse-grained). ST results are reported on the EN-CN test set.}
\label{tab:entropy-thresholds}
\begin{tabular}{lcccccccc}
\toprule
Threshold (Description) & Token Rate (Hz) & Task & WER (\%) & CER (\%) & BLEU & Q-MOS & S-MOS & UTMOS \\
\midrule
\multirow{3}{*}{Fine-grained}           
    & \multirow{3}{*}{24} & ASR & 6.4  & 3.5 & -    & -    & -    & - \\
    &                     & VC  & \textbf{6.7}  & \textbf{3.7} & -    & \textbf{4.10} & \textbf{3.95} & \textbf{3.98} \\
    &                     & ST  & - & -   & 30.1 & -    & -    & - \\
\midrule
\multirow{3}{*}{Moderately compressed}  
    & \multirow{3}{*}{15} & ASR & \textbf{5.6}  & \textbf{2.9} & -    & -    & -    & - \\
    &                     & VC  & 7.3  & 4.2 & -    & 3.82 & 3.76 & 3.80 \\
    &                     & ST  & - & -   & \textbf{31.5} & -    & -    & - \\
\midrule
\multirow{3}{*}{Coarse-grained}         
    & \multirow{3}{*}{7}  & ASR & 6.8  & 3.6 & -    & -    & -    & - \\
    &                     & VC  & 7.8  & 4.5 & -    & 3.68 & 3.58 & 3.61 \\
    &                     & ST  & - & -   & 31.0 & -    & -    & - \\
\bottomrule
\end{tabular}
\end{table*}

\subsection{Evaluation Metrics}

We evaluate ASR and voice conversion using both objective and subjective metrics. ASR performance is assessed by Word Error Rate (WER) and Character Error Rate (CER), while voice conversion quality is evaluated via Q-MOS (naturalness and clarity) and S-MOS (speaker similarity) through human ratings on Amazon Mechanical Turk, complemented by automatic predictions using UTMOS~\cite{utmos}. For speech-to-text translation, we report BLEU scores computed between the generated and reference text. Inference latency is measured as the average decoding time per input utterance, evaluated on a single NVIDIA V100 GPU over the test set.

\section{Exploration}

We conduct a series of experiments to examine how entropy-based token compression and the granularity of the compressed representations affect performance across various speech-related tasks. Our study centers on two key aspects:

\textbf{(1) Entropy-Based Semantic Compression with Varying Granularity.} We systematically explore how different entropy thresholds and boundary selection strategies, applied to the entropy-based language model, influence the granularity of compressed semantic representations. By modulating these parameters, we obtain semantic units of varying granularity. These compressed representations are then evaluated across multiple speech understanding and generation tasks, including ASR, speech translation, and voice conversion, offering empirical insights into the selection of optimal compression levels tailored to the specific characteristics and demands of each task.

\textbf{(2) Evaluation of Semantic Token Compression Methods.} We investigate the performance of four distinct semantic token compression approaches across multiple downstream tasks, providing a comprehensive assessment. These approaches include (i) original HuBERT tokens, (ii) deduplicated HuBERT tokens, (iii) fixed-length downsampled pooling tokens, and (iv) entropy-based compressed tokens at varying compression levels. This comparison highlights the advantages of our entropy-based compression method relative to alternative strategies.

\subsection{Experiment on Granularity Modulation Using Entropy Thresholds}

This experiment investigates the effect of varying entropy thresholds on the granularity of semantic tokens and their consequent impact on performance in ASR, speech-to-text translation (ST), and voice conversion (VC) tasks. The entropy threshold regulates the segmentation process and the extent of compression applied to the speech representations, resulting in token sequences with varying granularity levels. Specifically, three entropy thresholds are examined, yielding token compression rates of approximately 24 Hz for fine-grained tokens, 15 Hz for moderately compressed tokens, and 7 Hz for coarse-grained tokens. The performance metrics for each task under different entropy thresholds are summarized in Table~\ref{tab:entropy-thresholds}.

For ASR and speech-to-text translation tasks, moderate compression under the medium entropy threshold yields the best performance among the evaluated settings, as it aligns well with typical phoneme rates of around 17–18 Hz. Fine-grained tokens provide more detailed phonetic representations but may introduce redundancy and increase computational overhead. In contrast, coarse-grained tokens reduce token density at the cost of losing certain useful linguistic cues, which compromises recognition performance relative to the moderately compressed configuration.

Conversely, for the generative voice conversion task, semantic tokens that avoid being overly coarse in granularity, such as token rates below 10 Hz, tend to yield better performance in Q-MOS, S-MOS, and UTMOS. While coarser token representations may slightly improve S-MOS, they often reduce Q-MOS and intelligibility. Therefore, maintaining sufficiently fine-grained semantic token representations is important to capture the subtle acoustic nuances necessary for high-quality voice conversion. These findings underscore the necessity of adapting entropy-based compression parameters to the specific requirements of downstream tasks. Recognition-oriented applications such as ASR and ST benefit from a balanced compression strategy, whereas generative tasks like voice conversion require finer semantic granularity to ensure fidelity and naturalness in speech synthesis.

\begin{table}[h]
\small
\centering
\caption{ASR performance at 15 Hz token rate under different entropy-based segmentation criteria: M1 global threshold, M2 relative change, M3 combined.}
\label{tab:entropy-grouping}
\begin{tabular}{lcc}
\toprule
Method & WER (\%) & CER (\%) \\
\midrule
M1 & \textbf{5.8} & \textbf{3.1} \\
M2 & 6.3          & 3.4          \\
M3 & 5.9          & 3.2          \\
\bottomrule
\end{tabular}
\end{table}
\vspace{-1.0em}

\section{Entropy-Based Grouping Criteria}

We investigate the impact of two entropy-based grouping criteria on ASR performance under moderate token compression (15 Hz). The first method employs a global entropy threshold $\theta_g$ to identify tokens with high uncertainty as group boundaries. The second method detects abrupt changes in entropy by examining the relative difference between consecutive tokens, controlled by a relative threshold $\theta_r$. Additionally, we test a combined approach that applies both criteria simultaneously. As shown in Table~\ref{tab:entropy-grouping}, Method 1, which uses the global entropy threshold alone, achieves the best ASR performance, with a WER of 5.8\% and a CER of 3.1\%. Method 2, relying solely on relative entropy changes, results in slightly degraded accuracy. The combined approach provides marginal improvement over Method 2 but does not surpass the effectiveness of Method 1. These results suggest that the global entropy threshold is sufficient to segment semantic tokens into meaningful groups for ASR under moderate compression. Consequently, we adopt Method 1 as the default grouping strategy in subsequent experiments to maintain a balance between simplicity and performance.

\subsection{ASR Performance}

We evaluate the impact of various token compression strategies on ASR performance, measured by WER and CER. As shown in Table~\ref{tab:asr-results}, the \textbf{Entropy-guided (medium)} configuration (15 Hz), which approximates the phoneme rate, achieves the best recognition performance, slightly outperforming both \textbf{HuBERT (deduplicated)} (26 Hz) and \textbf{HuBERT (original)} (50 Hz). This suggests that moderate entropy-based compression effectively preserves essential linguistic information while reducing redundancy. The \textbf{Fixed-length pooling} approach with a window size of 2 (25 Hz) yields comparable results, but increasing the window to 4 (12.5 Hz) leads to a notable drop in accuracy, likely due to excessive smoothing that suppresses important fine-grained speech content details. Notably, the \textbf{Entropy-guided (high comp.)} setting (7 Hz), despite its lower token rate, the entropy-guided method outperforms the fixed-length pooling strategy at 12.5 Hz, demonstrating the effectiveness of entropy-based compression in preserving essential linguistic representations under substantial compression.

\begin{table}[h]    
\scriptsize
\centering
\caption{ASR performance under different token compression strategies.}
\label{tab:asr-results}
\begin{tabular}{lcc}
\toprule
\textbf{Method} & \textbf{WER (\%)} & \textbf{CER (\%)} \\
\midrule
HuBERT (original, 50Hz)          & 6.2 & 3.3 \\
HuBERT (deduplicated, 26Hz)      & 5.9 & 3.1 \\
Entropy-guided (medium, 15Hz)    & \textbf{5.6} & \textbf{2.9} \\
Fixed-length Pooling (2, 25Hz)   & 6.4 & 3.5 \\
Entropy-guided (high, 7Hz)       & 10.5 & 6.2 \\
Fixed-length Pooling (4, 12.5Hz) & 6.8 & 3.6 \\
\bottomrule
\end{tabular}
\end{table}
\vspace{-1.0em}

\subsection{Speech-to-Text Translation}

Table~\ref{tab:st-results} shows that Entropy-guided compression at 15 Hz achieves the best BLEU scores across all language pairs, indicating effective preservation of semantic content. The high compression variant (7 Hz) performs worse but still outperforms HuBERT deduplicated (26 Hz), suggesting that entropy-based selection better maintains key linguistic units under stronger compression. In contrast, Fixed-length Pooling leads to significant BLEU degradation, especially with larger pooling windows, due to the loss of important semantic and contextual details. Compared to ASR, speech-to-text translation is less dependent on fine-grained acoustic or phonetic information. The relative robustness of ST performance under coarser token compression suggests that moderate to coarse granularity sufficiently preserves the linguistic content needed for accurate translation.

\begin{table}[h]
\tiny
\centering
\caption{BLEU scores for speech-to-text translation with different compression methods. Higher BLEU score indicates better translation quality.}
\label{tab:st-results}
\begin{tabular}{l|c|c|c}
\toprule
\textbf{Method}               & \textbf{EN-CN} & \textbf{EN-FR} & \textbf{EN-ES} \\
\midrule
HuBERT (original, 50 Hz)           & 28.2  & 28.5  & 29.3  \\
HuBERT (deduplicated, 26 Hz)       & 29.8  & 30.2  & 30.7  \\
Entropy-guided (high comp., 7 Hz)  & 31.0  & 30.3  & 30.8  \\
Entropy-guided (medium, 15 Hz)     & 31.5  & 30.9  & 31.6 \\
Fixed-length Pooling (win=2, 25 Hz)       & 27.3  & 26.5 & 28.0  \\
Fixed-length Pooling (win=4, 12.5 Hz)       & 25.4  & 24.1  & 25.6  \\
\bottomrule
\end{tabular}
\end{table}
\vspace{-1.0em}

\subsection{Voice Conversion}
Table~\ref{tab:vc-results} shows that performance degrades progressively with increased compression. The original 50 Hz and deduplicated 26 Hz tokens yield the best results, while excessive reduction especially via fixed-length pooling significantly harms naturalness and intelligibility. Entropy-guided compression at 15 Hz maintains competitive quality, suggesting it can moderately reduce token rates without severely affecting performance. These results indicate that voice conversion relies more heavily on detailed acoustic and prosodic information, and is less tolerant to coarse compression than tasks like speech translation. 

\begin{table}[h]
\tiny
\centering
\caption{Voice conversion results.}
\label{tab:vc-results}
\begin{tabular}{l|c|c|c}
\toprule
\textbf{Method} & \textbf{Q-MOS} & \textbf{S-MOS} & \textbf{WER (\%)} \\
\midrule
HuBERT (original, 50 Hz)            & 3.89 & 3.75 & 6.2 \\
HuBERT (deduplicated, 26 Hz)        & \textbf{4.12} & \textbf{3.95} & 6.7 \\
Entropy-guided (medium, 15 Hz)      & 3.85 & 3.78 & \textbf{6.4} \\
Entropy-guided (high comp., 7 Hz)   & 3.33 & 3.27 & 9.1 \\
Fixed-length Pooling (win=2, 25 Hz) & 3.61 & 3.52 & 7.4 \\
Fixed-length Pooling (win=4, 12.5 Hz) & 3.25 & 3.12 & 9.9 \\
\bottomrule
\end{tabular}
\end{table}
\vspace{-1.0em}

\subsection{Decoding Latency}

Decoding latency decreases as token compression increases, with high compression yielding the fastest inference but reduced accuracy. Moderate compression achieves a favorable balance, significantly lowering latency while maintaining competitive performance. This highlights the importance of selecting appropriate compression levels to optimize ASR systems for efficiency.

\begin{table}[h]
\small
\centering
\caption{ASR decoding latency (ms) under different token compression methods.}
\label{tab:latency}
\begin{tabular}{lc}
\toprule
Compression Method & Latency (ms) \\
\midrule
HuBERT (original, 50 Hz)             & 172 \\
HuBERT (deduplicated, 26 Hz)         & 148 \\
Fixed-length Pooling (win=2, 25 Hz)  & 103 \\
Entropy-guided (medium, 15 Hz)       & 120 \\
Entropy-guided (high compression, 7 Hz) & 89 \\
\bottomrule
\end{tabular}
\end{table}

\section{Conclusion}
This work presents an entropy-based dynamic aggregation framework for compressing semantic speech representations by leveraging predictive uncertainty from a lightweight autoregressive language model. The proposed method effectively reduces token sequence length while preserving critical semantic information, enabling controllable granularity and improving computational efficiency. Experimental results on ASR, speech-to-text translation, and voice conversion tasks demonstrate that the compressed representations achieve comparable or superior performance to dense token sequences. These findings highlight the potential of entropy-guided compression as a flexible and scalable approach for efficient semantic speech modeling in diverse downstream applications.


\section{Limitations and Future Work}

Although the proposed entropy-based semantic representation compression framework demonstrates superior performance in semantic understanding tasks, enabling adjustable granularity according to task requirements, it exhibits limitations in generation-related tasks. Specifically, the quality of generated outputs tends to degrade as the compression ratio monotonically increases, suggesting that this approach may be less suitable for compressing sequences of acoustic tokens. Furthermore, additional experiments on other types of semantic tokens are necessary to validate the generalizability and broader applicability of the method. Future work will focus on addressing these limitations and exploring extensions to enhance the effectiveness of entropy-guided compression across a wider range of speech representation tasks.

\bibliography{custom}

\appendix

\label{sec:appendix}
\section{Detailed Experiment Settings}
\label{details_exp}
\subsection{Details in Subjective Evaluation}
We randomly select 50 sentences from the test set and perform the subjective evaluation on Amazon Mechanical Turk (MTurk). Each generated audio has been listened to by at least 10 native listeners. For Q-MOS evaluations, the listeners are instructed to focus on assessing the audio quality and naturalness while disregarding any differences in styles (such as timbre, emotion, and pronunciation). Conversely, for S-MOS evaluations, the listeners are instructed to concentrate on evaluating the speaker similarity to the audio prompt, while disregarding differences in content or audio quality. For the Q-MOS, S-MOS evaluations, each listener is asked to rate different speech samples using a Likert scale ranging from 1 to 5. 

\label{details_model}
\section{Details of Models}
\label{details_model}
An open-source HiFi-GAN vocoder\footnote{\url{https://www.modelscope.cn/models/iic/CosyVoice-300M}} is used to synthesize speech from mel-spectrograms. In the cross-attentive local encoder, we employ 4 layers of Transformer blocks. To obtain semantic representations at different levels of granularity, we adjust the entropy threshold $\theta_g$ within the range of 7.5 to 10.5. Specifically, for token rates of 7 Hz, 15 Hz, and 24 Hz, the thresholds are set to $\theta_g = 9.7$, $8.5$, and $7.8$, respectively.

For the ASR task, we adopt an autoregressive decoder architecture similar to Whisper, utilizing an encoder-decoder structure with a total of approximately 104M parameters to generate text from compressed token sequences. For voice conversion (VC), we follow the design of Seed-VC \cite{liu2024zero} and use a Diffusion Transformer model as the generator. The compressed semantic tokens are aligned with the mel-spectrogram in time by padding zero tensors, and combined with mel-spectrogram features as additional conditions. These are then fed into a flow matching model to generate the final mel-spectrograms.

\label{Additional Experiments}
\section{Additional Experiments}
\label{Additional Experiments}
\subsection{About Cross-Attentive Local Encoder}

To further clarify the effectiveness of our \textbf{Cross-Attentive Local Encoder (CALE)} module, we conducted additional ablation studies across three tasks. The results are summarized below:

\begin{table}[h!]
\small
\tabcolsep=2.5pt
\centering
\caption{15Hz token rate for ASR experiment.}
\label{tab:asr_experiment}
\begin{tabular}{lcc}
\toprule
\textbf{Method} & \textbf{WER} & \textbf{CER} \\
\midrule
boundary + max pooling (w/o CALE) & 6.4 & 3.4 \\
boundary + average pooling (w/o CALE) & 6.8 & 3.6 \\
boundary + attention pooling (w/o CALE) & 6.1 & 3.3 \\
\textbf{Ours (with CALE)} & \textbf{5.6} & \textbf{2.9} \\
\bottomrule
\end{tabular}
\end{table}

\begin{table}[h!]
\scriptsize
\tabcolsep=2.5pt
\centering
\caption{24Hz token rate for voice conversion experiment.}
\label{tab:vc_experiment}
\begin{tabular}{lccc}
\toprule
\textbf{Method} & \textbf{Q-MOS} & \textbf{S-MOS} & \textbf{WER (\%)} \\
\midrule
boundary + max pooling (w/o CALE) & 4.02 & 3.88 & 7.1 \\
boundary + average pooling (w/o CALE) & 3.97 & 3.85 & 7.4 \\
boundary + attention pooling (w/o CALE) & 4.06 & 3.92 & 6.9 \\
\textbf{Ours (with CALE)} & \textbf{4.10} & \textbf{3.95} & \textbf{6.7} \\
\bottomrule
\end{tabular}
\end{table}

\begin{table}[h!]
\scriptsize
\tabcolsep=2.5pt
\centering
\caption{15Hz token rate for speech-to-text translation experiment. \textbf{CALE} refers to our Cross-Attentive Local Encoder.}
\label{tab:s2t_experiment}
\begin{tabular}{lccc}
\toprule
\textbf{Method} & \textbf{EN-CN} & \textbf{EN-FR} & \textbf{EN-ES} \\
\midrule
boundary + max pooling (w/o CALE) & 30.1 & 29.4 & 29.8 \\
boundary + average pooling (w/o CALE) & 29.2 & 28.5 & 29.1 \\
boundary + attention pooling (w/o CALE) & 30.8 & 30.2 & 30.7 \\
\textbf{Ours (with CALE)} & \textbf{31.5} & \textbf{30.9} & \textbf{31.6} \\
\bottomrule
\end{tabular}
\end{table}


\noindent From the above ablation study results, we can observe that when only using the corresponding grouping boundaries + pooling methods for aggregation without incorporating the Cross-Attentive Local Encoder, our method shows varying degrees of performance degradation across all three tasks. We analyze that beyond helping with intra-group information aggregation, the introduction of the Cross-Attentive Local Encoder module enables the model to achieve global perception of the original input, supplementing missing global information. This global perception capability allows the model to better understand the contextual relationships both within and across semantic boundaries, thereby enhancing overall performance.

\subsection{About the K-means clusters}

Due to computational resource and time constraints, we conducted additional experiments with HuBERT representations using K=2000 clusters to evaluate the performance of our compression framework across different token rates on ASR and ST tasks. The results are presented in the table \ref{tab:hubert_k2000_results}.

\begin{table*}[ht]
\centering
\caption{Performance of HuBERT representations (K=2000) on ASR and ST tasks across different token rates.}
\label{tab:hubert_k2000_results}
\begin{tabular}{lccccc}
\toprule
\textbf{Threshold (Description)} & \textbf{Token Rate (Hz)} & \textbf{Task} & \textbf{WER (\%)} & \textbf{CER (\%)} & \textbf{BLEU} \\
\midrule
\multirow{2}{*}{Fine-grained} & \multirow{2}{*}{24} & ASR & 4.5 & 2.7 & - \\
                               &                      & ST  & -   & -   & 31.7 \\
\midrule
\multirow{2}{*}{Moderately compressed} & \multirow{2}{*}{15} & ASR & 3.9 & 2.4 & - \\
                               &                      & ST  & -   & -   & 33.2 \\
\midrule
\multirow{2}{*}{Coarse-grained} & \multirow{2}{*}{7}  & ASR & 5.8 & 3.1 & - \\
                               &                      & ST  & -   & -   & 32.1 \\
\bottomrule
\end{tabular}
\end{table*}

Compared to our original experiments with K=500, the results with K=2000 demonstrate consistent improvements across both ASR and ST tasks. Most importantly, our core finding remains valid: the dynamic compression framework effectively identifies optimal compression rates for different tasks, with moderately compressed representations (15 Hz) achieving the best efficiency-performance trade-off.

Regarding alternative SSL features like WavLM, we acknowledge that both HuBERT and WavLM encode high-level semantic representations. We believe our compression framework would naturally transfer to these models, though resource constraints prevented comprehensive evaluation. These additional experiments with K=2000 strengthen our analysis while confirming the robustness of our proposed framework across different parameter settings.

\subsection{About semantic structure}
We also conducted boundary alignment analysis comparing our entropy-based compressed token boundaries with linguistically meaningful boundaries obtained through the open-source Montreal Forced Aligner (MFA) tool \footnote{\url{https://github.com/MontrealCorpusTools/Montreal-Forced-Aligner}} on the test-clean subset.

\begin{table}[h]
\scriptsize
\tabcolsep=2.5pt
\centering
\caption{Boundary alignment analysis at different compression rates. CR: Compression Rate; ASD: Avg Span Duration; WBA: Word Boundary Alignment; PBA: Phoneme Boundary Alignment; MTD: Mean Time Deviation.}
\label{tab:boundary_alignment}
\begin{tabular}{lccccc}
\hline
\textbf{CR (Hz)} & \textbf{ASD (ms)} & \textbf{WBA (\%)} & \textbf{PBA (\%)} & \textbf{MTD (ms)} \\ \hline
Original (50Hz)    & 20                & -                 & -                 & -                 \\
Deduplicated (26Hz) & 38                & 11.4              & 32.7              & 42.3              \\
24Hz               & 42                & 18.9              & 35.3              & 38.7              \\
15Hz               & 67                & 22.6              & 83.2              & 15.8              \\
7Hz                & 143               & 89.7              & 28.1              & 28.6              \\ \hline
\end{tabular}
\end{table}

\textbf{Metrics} : Word/Phoneme Boundary Alignment measures the percentage of compressed boundaries that fall within \textpm50ms of MFA-aligned boundaries. Mean Time Deviation represents the average absolute temporal distance between compressed and reference boundaries.

According to the table, our entropy-based compression method demonstrates clear frequency-boundary correspondence relationships. The 15Hz compression rate achieves optimal phoneme boundary alignment (83.2\%), with its 67ms average span duration closely matching phoneme-level linguistic units. The 7Hz high compression rate reaches the highest performance in word boundary alignment (89.7\%), with its 143ms span duration corresponding to typical word unit durations. The 26Hz deduplication processing, with its 38ms span duration falling between phoneme and word units, shows moderate performance on both boundary alignments. This systematic correspondence pattern demonstrates that our compressed tokens capture linguistically meaningful structures at different granularities, with "semantic structure" referring to the hierarchical linguistic units preserved at each compression level.

\subsection{About the fixed-length downsampling}

In our original experiments, the fixed-length downsampling refers to using a fixed-sized window to aggregate the representations within that window, where we applied max pooling as the default aggregation method. Since different pooling strategies may affect performance to some extent, we conducted additional ASR experiments with various pooling methods under the fixed-length setting. The results are shown in the table below:

\begin{table}[h]
\scriptsize
\tabcolsep=2.5pt
\centering
\caption{ASR results with various pooling methods under the fixed-length setting.}
\label{tab:pooling_comparison}
\begin{tabular}{lcc}
\hline
\textbf{Method}                               & \textbf{WER (\%)} & \textbf{CER (\%)} \\ \hline
Entropy-guided (medium, 15Hz)                 & 5.6               & 2.9               \\
Fixed-length max pooling (2, 25Hz)            & 6.4               & 3.5               \\
Fixed-length average pooling (2, 25Hz)        & 6.7               & 3.8               \\
Fixed-length attention pooling (2, 25Hz)      & 6.2               & 3.4               \\
Fixed-length max pooling (4, 12.5Hz)          & 6.8               & 3.6               \\
Fixed-length average pooling (4, 12.5Hz)      & 7.2               & 3.9               \\
Fixed-length attention pooling (4, 12.5Hz)    & 6.6               & 3.5               \\ \hline
\end{tabular}
\end{table}

The results show that while different pooling strategies under fixed-length settings lead to slightly varied performance, none surpass our entropy-guided method. This confirms that the gain comes from adaptive boundary modeling rather than pooling choice alone. The performance gaps also become more pronounced at lower token rates (e.g., 12.5Hz), highlighting the importance of boundary quality.

\subsection{Ablation on Entropy Thresholds}
In Table \ref{tab:entropy-thresholds}, we present the performance of Boundary Identification Method 1 (global entropy threshold $\theta_g$) across three tasks at different compression granularities: 24Hz with $\theta_g = 0.78$, 15Hz with $\theta_g = 0.85$, and 7Hz with $\theta_g = 0.97$. In Table 2, we focus on 15Hz (the optimal token rate for ASR) and compare three segmentation strategies: $\theta_g$ only (M1), $\theta_r$ only (M2), and Combined ($\theta_g + \theta_r$, M3). Results show that M1 and M3 consistently outperform M2, and their performance is close. To keep the experimental setup simple and efficient, we adopt the $\theta_g$-only method in the main experiments for boundary identification. Internally, we also tested the $\theta_g$-only approach across a wider range of token rates (6Hz–30Hz) on all three tasks and selected 24Hz, 15Hz, and 7Hz as representative settings.

We acknowledge that further exploration of $\theta_g$ and $\theta_r$ combinations is meaningful, but it requires substantial computational resources. Based on Table \ref{tab:entropy-grouping}, we now provide extended results comparing all three methods (M1, M2, M3) under different granularities on ASR (WER $\downarrow$) and ST (BLEU, EN-CN $\uparrow$):

\begin{table}[h]
\scriptsize
\tabcolsep=2.5pt
\centering
\caption{Comparison of boundary identification methods across different granularities. Note: M1 = global threshold ($\theta_g$), M2 = relative change ($\theta_r$), M3 = combined ($\theta_g + \theta_r$). For metrics, $\downarrow$ indicates lower is better, and $\uparrow$ indicates higher is better.}
\label{tab:method_comparison}
\begin{tabular}{llcc}
\hline
\textbf{Token Rate} & \textbf{Method}                          & \textbf{ASR (WER $\downarrow$)} & \textbf{ST (BLEU $\uparrow$)} \\ \hline
\multirow{3}{*}{24Hz} & M1 ($\theta_g =0.78$)                  & 6.4                             & 30.1                          \\
                    & M2 ($\theta_r =0.65$)                  & 7.0                             & 29.4                          \\
                    & M3 ($\theta_g =0.75, \theta_r =0.62$)    & 6.3                             & 30.3                          \\ \hline
\multirow{3}{*}{15Hz} & M1 ($\theta_g =0.85$)                  & 5.8                             & 31.5                          \\
                    & M2 ($\theta_r =0.69$)                  & 6.3                             & 30.2                          \\
                    & M3 ($\theta_g =0.82, \theta_r =0.65$)    & 5.9                             & 31.2                          \\ \hline
\multirow{3}{*}{7Hz}  & M1 ($\theta_g =0.97$)                  & 6.8                             & 31.0                          \\
                    & M2 ($\theta_r =0.71$)                  & 7.5                             & 29.8                          \\
                    & M3 ($\theta_g =0.86, \theta_r =0.68$)    & 7.0                             & 31.2                          \\ \hline
\end{tabular}
\end{table}

Observed the above table, we can draw several important conclusions that extend our findings from Table \ref{tab:entropy-grouping}. While our previous analysis focused on the 15Hz setting (optimal for ASR tasks), this comprehensive evaluation across multiple token rates (24Hz, 15Hz, 7Hz) provides stronger evidence for our boundary identification approach. The consistent performance pattern M1$\approx$M3$>$M2 across different compression granularities reinforces our decision to adopt the single global threshold $\theta_g$ method for subsequent experiments. Notably, the performance gaps remain relatively stable across token rates, with M1 and M3 consistently achieving lower WER (5.8-6.8) and higher BLEU scores (30.1-31.5) compared to M2, which demonstrates that the segmentation methods of M1 and M3 are more reasonable. Through more extensive ablation studies on $\theta_g$ and $\theta_r$ combinations, we observe that the global entropy threshold is sufficient to segment semantic tokens into meaningful groups.



\end{document}